\documentclass[lettersize,journal]{IEEEtran}
\usepackage{amsmath,amsfonts}
\usepackage{algorithmic}
\usepackage{algorithm}
\usepackage{array}
\usepackage{textcomp}
\usepackage{stfloats}
\usepackage{url}
\usepackage{soul}
\usepackage{verbatim}
\usepackage{graphicx}
\usepackage{cite}
\usepackage{tikz}
\usepackage{bm}
\usetikzlibrary{positioning}
\usepackage{overpic}
\usepackage{comment}
\usepackage{amsmath,amssymb} 
\usepackage{color}
\usepackage{animate}
\usepackage{makecell}
\usepackage{multirow}
\usepackage{booktabs}
\usepackage{sidecap}
\usepackage{import}
\usepackage{verbatim}
\usepackage{graphicx}
\usepackage{cite}
\usepackage{paralist, tabularx}
\usepackage{paralist}
\usepackage{graphicx, animate}
\usepackage{subfigure}
\usepackage[citecolor=blue, colorlinks]{hyperref}

\begin{document}

\title{Decoupling Dynamic Monocular Videos for Dynamic View Synthesis}

\author{Meng You and Junhui Hou
\thanks{This work was supported in part by Hong Kong Research Grants Council
under Grant 11218121, and in part by Hong Kong Innovation and Technology
Fund under Grant MHP/117/21.}
\thanks{The authors are with the Department of Computer Science, City University of Hong Kong, Hong Kong SAR. Email: mengyou2-c@my.cityu.edu.hk; jh.hou@cityu.edu.hk}}

\markboth{ }%
{Shell \MakeLowercase{\textit{et al.}}: A Sample Article Using IEEEtran.cls for IEEE Journals}

\maketitle

\begin{abstract}
The challenge of dynamic view synthesis from dynamic monocular videos, i.e., synthesizing novel views for free viewpoints given a monocular video of a dynamic scene captured by a moving camera, 
mainly lies in accurately modeling the \textbf{dynamic objects} of a scene using limited 2D frames, each with a varying timestamp and viewpoint. Existing methods usually require pre-processed 2D optical flow and depth maps by off-the-shelf methods to supervise the network, making them suffer from the inaccuracy of the pre-processed supervision and the ambiguity when lifting the 2D information to 3D. 
In this paper, we tackle this challenge in an unsupervised fashion. 
Specifically, we decouple the motion of the dynamic objects into object motion and camera motion, 
respectively regularized by proposed unsupervised surface consistency and patch-based multi-view constraints.
The former enforces the 3D geometric surfaces of moving objects to be consistent over time, while the latter regularizes their appearances  to be consistent across different viewpoints. Such a fine-grained motion formulation can alleviate the learning difficulty for the network, thus enabling it to produce not only novel views with higher quality but also more accurate scene flows and depth than existing methods requiring extra supervision. We will make the code publicly available at \url{https://github.com/mengyou2/DecoulpingNeRF}.
\end{abstract}


\section{Introduction}

\begin{figure*}[t]
\hsize=1\textwidth 
    \centering
    \subfigure[Input Video]{
    \centering
    \animategraphics[width=0.392\textwidth, loop, autoplay]{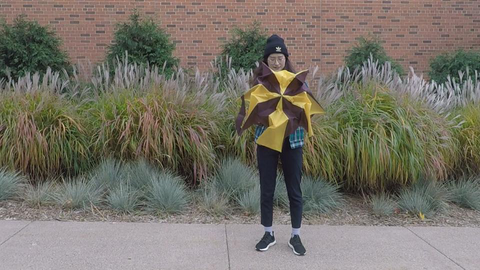}{demo/umbrella/}{0}{11}
    }
    \subfigure[Gao \textit{et al.} \cite{gao2021dynamic}]{
    \centering
    \animategraphics[width=0.268\textwidth, loop, autoplay]{12}{demo/umbrella_crop/gao_}{1}{88}
    }
    \subfigure[Ours]{
    \centering
    \animategraphics[width=0.268\textwidth, loop, autoplay]{12}{demo/umbrella_crop/ours_}{1}{88}
    }\vspace{-0.3cm}
    \caption{Given a monocular video of a dynamic scene captured by a moving camera, rendered novel views (zoomed in at dynamic part) at arbitrary viewpoints and any input timestep by state-of-the-art method Gao \textit{et al.} \cite{gao2021dynamic} and Ours. Please use Adobe Acrobat to display these videos.}
    \label{figure: setting}
\end{figure*}

Novel view synthesis aims to generate new views of a scene from existing views or 3D models,
allowing for immersive experiences in virtual/augmented reality \cite{wei2019vr,collet2015high}, robotics \cite{manuelli2019kpam}, and so on. In a more general scenario, dynamic view synthesis from dynamic monocular videos makes this field more prominent, in which given a monocular video of a scene containing moving objects captured by a moving camera, novel views at arbitrary viewpoints and timestamps will be synthesized to create visually stunning effects like the bullet-time effect, as demonstrated in Fig. \ref{figure: setting}.
However,  dynamic view synthesis is much more challenging due to the difficulty of accurately modeling dynamic objects in the scene using limited 2D frames, each with a varying timestamp and viewpoint.

\begin{figure}[t]
    \centering
    \includegraphics[width=0.48\textwidth]{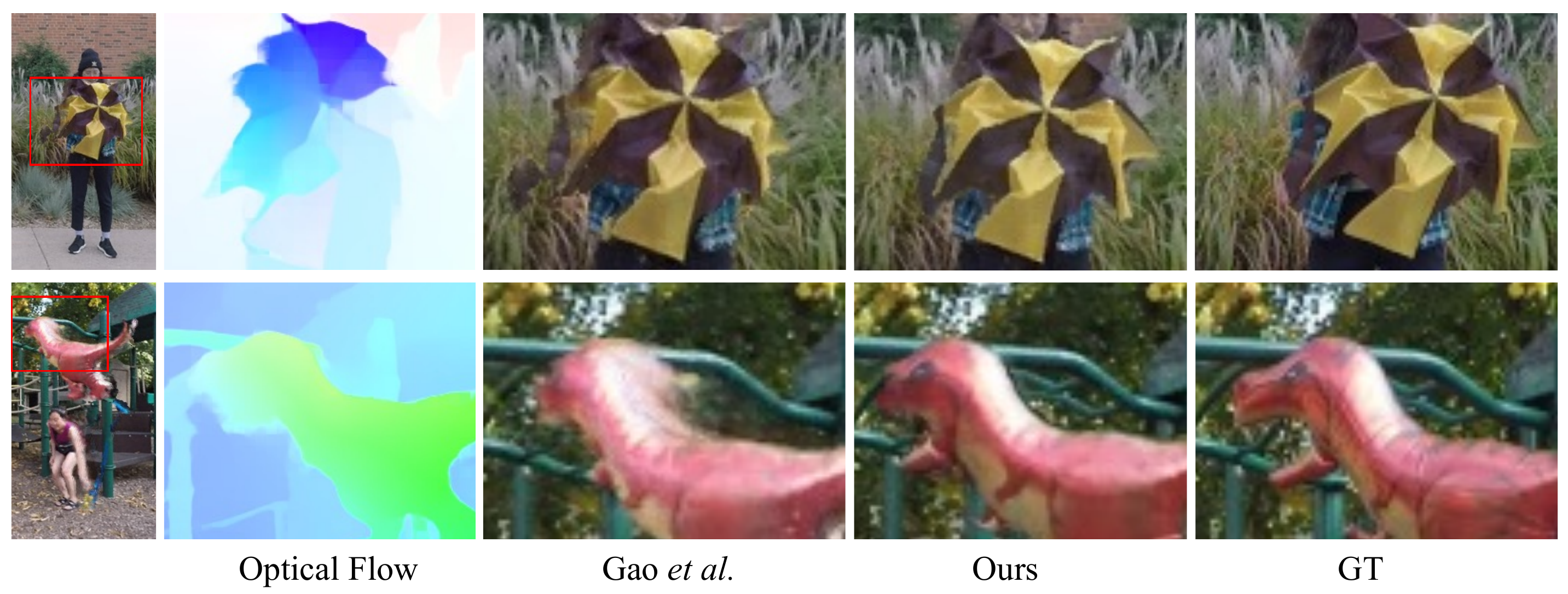}\vspace{-0.3cm}
    \caption{Visual illustration of the limitation of existing method Gao \textit{et al}. \cite{gao2021dynamic}. The pre-processed 2D optical flow maps used for training networks contain severe errors, resulting in the corresponding parts of synthesized novel views being wrong.}
    \label{fig:drawbacks} 
\end{figure}

Recently, the advancements in deep learning have led to remarkable achievements in novel view synthesis. One of the most notable works in this field is the neural radiance field (NeRF) \cite{mildenhall2020nerf}, which represents a scene using a continuous 5D volumetric function encoded by a multi-layer perceptron. 
Efforts are being made to expand the use of NeRF to dynamic view synthesis, 
such as learning a deformable warp field between different time frames and a canonical space \cite{pumarola2021d,tretschk2021non,park2021nerfies}  
 or a neural scene flow field (NSFF) between adjacent frames \cite{li2021neural,gao2021dynamic,du2021neural,wang2021neural}. Moreover, the NSFF is more capable of maintaining continuity across timestamps in real-world monocular videos than the deformable NeRF. 
 However, previous methods 
 usually require pre-processed 2D optical flow and depth maps by additional methods as supervision during network training, making them suffer from the following limitations.
First, the inevitable local errors of the pre-processed supervision would harm the network and negatively affect the reconstruction quality, as demonstrated in Fig. \ref{fig:drawbacks}. Second, using 2D maps to supervise 3D geometry would introduce ambiguity and result in inaccurate reconstructions. Finally,  obtaining accurate pre-processed maps may be computationally expensive and time-consuming, making it inconvenient.

In contrast to existing methods, we aim to tackle this challenge in an \textit{unsupervised} manner, i.e., without using pre-processed information by off-the-shelf methods as supervision. 
Specifically, we propose to decouple the motion of dynamic objects in a scene  
into object motion and camera motion.  
Then, we propose 
two novel unsupervised regularization terms named surface consistency and multi-view consistency for modeling object motion and camera motion, respectively. 
The former ensures that the surface of moving objects remains geometrically consistent over time, while the latter ensures consistent appearance across different viewpoints. 
Such a fine-grained formulation of the motion of dynamic objects makes it easier for the network to model them. 
\textit{It is impressive that our unsupervised method
even substantially improves  state-of-the-art supervised methods. Besides, our method can output more accurate scene flow and depth. We believe our method will provide a promising baseline for dynamic view synthesis.}

In summary, the main contributions of this paper are:  
\begin{compactitem} 
    \item  fine-grained formulation of the motion of dynamic objects in a scene in a decoupling fashion;  
    \item unsupervised surface consistency regularization for modeling object motion; 
    \item unsupervised patch-based multi-view regularization for modeling camera motion; and 
    \item current state-of-the-art performance of dynamic view synthesis and unsupervised scene flow and depth prediction.
\end{compactitem}

The rest of this paper is orgarnized as follows. Section \ref{sec: related work}
briefly reviews related works. Section \ref{sec: prelimiary} reviews the general pipeline of NSFF-based novel view synthesis for dynamic videos.
Section \ref{sec:method} presents the proposed
constraints for optimizing dynamic NeRF, followed by comprehensive
experiments and analyses in Section \ref{sec: experiments}. Finally, Section \ref{sec: conclusion and discussion}
concludes the paper and discusses some future directions for
improving the proposed method.
\section{Related Work}
\label{sec: related work}
\subsection{Novel View Synthesis for Static Scenes}
View synthesis for static scenes is a long-standing problem in computer vision/graphics. Owing to the development of deep learning techniques, recent image-based rendering learns scene information with neural networks.   
Some methods \cite{zhou2016view,chen2019monocular,sun2018multi,park2017transformation} adopt neural networks to estimate depth or appearance flow explicitly between viewpoints and use it to warp pixels from sources to the novel view. 
Many recent methods choose to learn 3D scene geometry with explicit scene representations which can be rendered to novel viewpoints.  
 Some \cite{yan2016perspective,choy20163d,jimenez2016unsupervised,kar2017learning,penner2017soft,xie2019pix2vox,henzler2019escaping,lombardi2019neural} learn voxel-based grid from source images, representing objects as a 3D volume in the form of voxel occupancies.
Another class of methods \cite{zhou2018stereo,flynn2019deepview,mildenhall2019local,srinivasan2019pushing,li2020crowdsampling} use multiplane images (MPI) to represent the scene as a set of front-parallel planes at fixed depths, where each plane consists of an RGB image and an $\alpha$ map.  
Similar to MPI, LDI \cite{shade1998layered,swirski2011layered,tulsiani2018layer,shih20203d,dhamo2019object} keeps several depth and color values at every pixel and renders images by a back-to-front forward warping algorithm. 
Another explicit scene representation is the 3D point cloud. Some methods \cite{wiles2020synsin,le2020novel,cao2022fwd} project 2D images or features to the 3D point cloud space and generate novel views with a differentiable renderer.

More recent methods represent the scene as an NeRF \cite{mildenhall2020nerf} by learning a continuous volumetric scene function. 
Following methods extend NeRF in different aspects: reconstruction quality \cite{zhang2020nerf++,barron2021mip}, generalization \cite{yu2021pixelnerf,Wang_2021_CVPR,trevithick2020grf,SRF,li2021mine,chen2021mvsnerf}, unconstrained scene \cite{martin2021nerf,barron2022mip}, and so on.\\

\subsection{Novel View Synthesis for Dynamic Scenes}
Novel view synthesis for dynamic scenes can be more challenging because multi-view geometry consistency does not apply to contents with motion. Recently, some image-based rendering methods \cite{yoon2020novel,lin2021deep} have achieved impressive performance for dynamic view synthesis. Yoon \textit{et al.} \cite{yoon2020novel} proposed to blend depth-based warping with blending techniques for dynamic view synthesis. They used multi-view stereo for static parts and monocular depth estimation for dynamic parts, which are then blended for warping to the final result. 
Lin \textit{et al.} \cite{lin2021deep} modeled a 3D scene using deep 3D mask volume representation based on MPI. The 3D scene is modeled as a background and an instantaneous MPI, added to a global MPI with predicted masks. However, these image-based rendering methods rely heavily on accurate depth estimation and are not flexible enough to render novel views from arbitrary viewpoints.

Recently, NeRF has achieved impressive performance in view synthesis tasks. Many methods have extended NeRF to address dynamic view synthesis. 
Some recent methods \cite{weng2022humannerf,gafni2021dynamic,peng2021animatable,noguchi2021neural} focus on specific domains, such as the face or human synthesis. Others focus on more general settings with video inputs capturing a global scene containing static and dynamic objects. There are three main ideas for extending NeRF: (1) incorporating time information or time-related latent codes into the input of NeRF \cite{xian2021space,li2022neural}; (2) learning a deformable NeRF \cite{pumarola2021d,tretschk2021non,park2021nerfies}; and (3) learning 3D neural flow fields \cite{li2021neural,gao2021dynamic,du2021neural,wang2021neural}.

Deformable NeRF splits the rendering process into two parts: a base NeRF encoding the canonical space and a warp field that maps the canonical NeRF into a deformed scene at another time.
A multi-layer perceptron (MLP) is used to encode the frames at $t=0$, which is considered the canonical scene, while another MLP takes the points' positions and time as input to produce the displacement between the point's position in the current time and the canonical space. Pumarola \textit{et al}. \cite{pumarola2021d} applied this approach to dynamic synthetic objects. Tretschk \textit{et al.} \cite{tretschk2021non} further extended it to real dynamic scenes from monocular video by incorporating a rigidity prediction network. Park \textit{et al}. introduced an elastic regularization of the deformation field to improve the regularization of NeRFs \cite{park2021nerfies} and further handled topological changes by lifting NeRFs into a higher dimensional space \cite{park2021hypernerf}. Based on \cite{park2021hypernerf},   Wu \textit{et al}. \cite{wu2022d} proposed to segment dynamic objects from the static background in a monocular video by using separate neural radiance fields to represent the moving objects and the static background in a self-supervised manner.
However, the main challenge in deformable NeRF is handling large motions in the scene, as all frames are warped from the same canonical space, resulting in a lack of continuity throughout the sequence.

Neural scene flow fields model the dynamic scene as a time-variant continuous function of appearance, geometry, and 3D scene motion. 
Li \textit{et al}. \cite{li2021neural} proposed to estimate 3D motion by modeling scene flow as a dense 3D vector field. Given a 3D point and time, their methods output not only reflectance and opacity but also forward and backward scene flow.
Gao \textit{et al}. \cite{gao2021dynamic} adopted the same formulation and improved the reconstruction quality by further regulating the temporal consistency and rigidity of the neural scene flow. 
Du \textit{et al}. \cite{du2021neural} addressed a set of consistency losses regarding scene appearance, density, and motion to enforce consistency through the flow fields.
Wang \textit{et al}. \cite{wang2021neural} adopt a neural trajectory field to model the scene flow and parameterize the trajectory as a combination of sinusoidal bases, i.e., the discrete cosine transform. 
Li \textit{et al}. \cite{li2023dynibar} adopted a volumetric image-based rendering framework that addresses the challenges of blurry or inaccurate renderings in long videos for dynamic view synthesis by aggregating features from nearby views. Tian \textit{et al.} \cite{23iccv/tian_mononerf} introduced a method aimed at jointly learning point features and scene flows, incorporating constraints such as point trajectory and feature correspondence across frames.
Despite inheriting the simplicity and efficiency of NeRF, these methods often require additional regulations, particularly preprocessing data-driven terms, to function optimally. 

Recently, the domain of neural rendering has witnessed groundbreaking advancements with the advent of 3D Gaussian Splatting (3D-GS) \cite{kerbl20233d}. This innovative technique employs a specialized CUDA-based differentiable Gaussian rasterization pipeline, markedly accelerating rendering speeds. Building upon the capabilities of 3D-GS, several studies have ventured into applying it for dynamic view synthesis. Wu \textit{et al}. \cite{wu20234d} developed 4D Gaussian Splatting (4D-GS), which synergizes 3D-GS with 4D neural voxels sourced from HexPlane. This method adeptly crafts Gaussian features from 4D neural voxels, subsequently utilizing MLPs to anticipate Gaussian deformations at novel time points. Yang \textit{et al}. \cite{yang2023deformable} integrated a deformable framework into the 3D-GS. They facilitated scene reconstruction using 3D-GS and employed a deformation field to precisely depict dynamic scenes. Li~\textit{et al}. \cite{li2023spacetime}  utilized a collection of Spacetime Gaussians (STG) to characterize dynamic scenes, augmenting each STG with temporal opacity, polynomial motion/rotation, and time-dependent attributes. Luiten ~\textit{et al.} \cite{luiten2023dynamic} conceptualized dynamic scenes by permitting Gaussians to undergo motion and rotation across time, while ensuring their color, opacity, and size remain consistent.


\begin{figure}[t]
\centering
\includegraphics[width=0.47\textwidth]{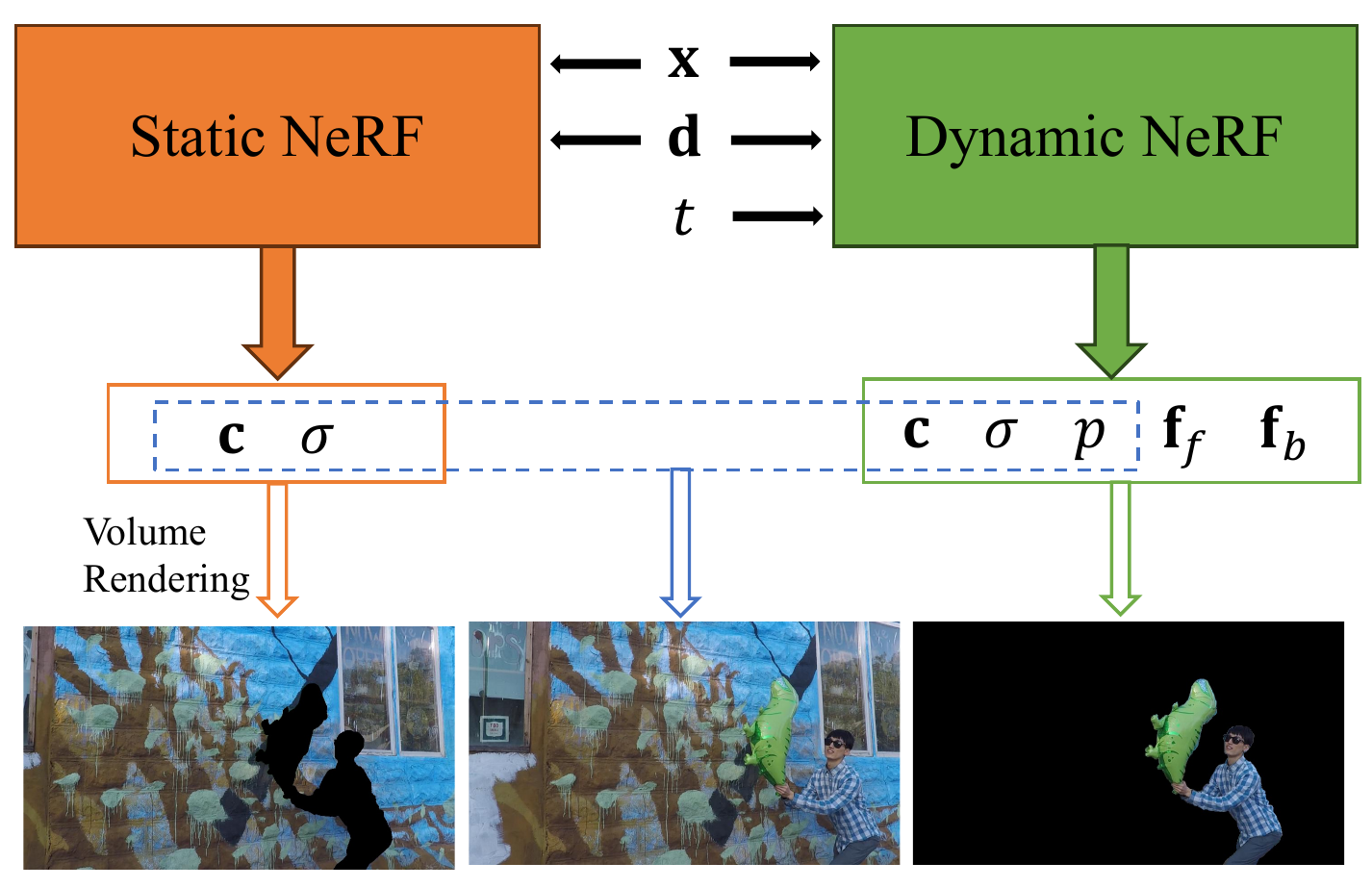}\vspace{-0.4cm}
\caption{ Our approach follows existing methods by utilizing two neural radiance fields: a \textit{static NeRF} \cite{mildenhall2020nerf} for modeling the static background of a scene and an \textit{NSFF} \cite{li2021neural,gao2021dynamic} for modeling the dynamic objects and the scene flow. We train a Static NeRF model to reconstruct the scene background, excluding pixels marked as dynamic during training. We train a Dynamic NeRF that takes (x, d) and t as input for modeling dynamic objects. This allows the Dynamic NeRF to predict color $\textbf{c}$, density $\sigma$, 3D scene flow $\textbf{f}_f, \textbf{f}_b$, and blending weight $p$. The blending weight $p$ enables the composition of static and dynamic NeRF components.
}
\label{figure: decoupling}
\end{figure}

\section{Preliminary}
\label{sec: prelimiary}
Let $\mathcal{I}:=\left\{\mathbf{I}_{n}\in\mathbb{R}^{H\times W}\right\}_{n=1}^N$ denote a dynamic monocular video, consisting of $N$ frames, each of dimensions $H\times W$. Each frame $\mathbf{I}_n$ associates with a timestamp $t_n$ and a camera pose $\mathbf{p}_n$. 
Given  $\mathcal{I}$ and $\{t_n, \mathbf{p}_n\}_{n=1}^N$, 
dynamic view synthesis aims at synthesizing novel views at arbitrarily given viewpoints and timestamps. 

This task is challenging because 
the 3D geometry of dynamic objects in the scene 
varies across different timestamps, while the monocular video only provides a single 2D observation of them 
at each timestamp.
Existing methods \cite{li2021neural,gao2021dynamic} use two neural radiance fields: a \textit{static NeRF} \cite{mildenhall2020nerf} to model the static background of a scene, and an \textit{NSFF} 
\cite{li2021neural,gao2021dynamic} to model the dynamic objects and the scene flow.\\

\noindent\textbf{Static NeRF} adopts the plain 
NeRF to model the static background of a scene. Specifically, for a 3D point $\mathbf{x}\in \mathbb{R}^3$ sampled on a ray $\mathbf{r}(s) = \mathbf{o}+s\mathbf{d}$ with the 
origin $\mathbf{o}\in \mathbb{R}^3$ and the normalized viewing direction $\mathbf{d} \in \mathbb{R}^3$, the static NeRF uses an MLP $\texttt{MLP}_s(\cdot)$ to map directional $\mathbf{x}$ to an RGB color $\mathbf{c}_s $  and a density $\sigma_s$: 
\begin{equation}
    (\mathbf{c}_s,\sigma_s)= \texttt{MLP}_s(\mathbf{x},\mathbf{d}).
\end{equation} 
The static NeRF is trained  by minimizing
\begin{equation}
    \mathcal{L}_{static} = \sum_{\mathbf{r}\in \mathcal{R} }  \| (\mathbf{\widehat{C}}_s(\mathbf{r})- \mathbf{C}(\mathbf{r}))(1-M(\mathbf{r})) \|_{2}^{2}, 
\end{equation}
where $\mathcal{R}$ is the set of rays in each batch,  $\mathbf{\widehat{C}}_s(\mathbf{r})$ computed with the volume rendering integral is the rendered color of the pixel the ray $\textbf{r}$ passes through, $\mathbf{C}(\mathbf{r})$ is the ground truth color value, and $M$ is the preprocessed mask indicating the dynamic part.\\

\noindent\textbf{NSFF.} 
Taking $(\mathbf{x},~\mathbf{d},~t)$ as input, where $t$ is the timestamp of the frame, NSFF outputs the forward 3D scene flow $\mathbf{f}_{f}$ from $t$ to $t+1$, the backward 3D scene flow $\mathbf{f}_{b}$ from $t$ to $t-1$, and a probability value $p$ indicating the dynamic part, in addition to the volume density and view-dependent emitted radiance: 
\begin{equation}
    (\mathbf{c}^t_d,\sigma^t_d,\mathbf{f}_{f},\mathbf{f}_{b},p)=\texttt{MLP}_d(\mathbf{x},\mathbf{d},t),
\label{dynamicnerf}
\end{equation}
where $\texttt{MLP}_d(\cdot)$ is the learnable MLP. 

Thus, time neighboring points can be computed easily by $\mathbf{x}+ \mathbf{f}_{f}$ and $\mathbf{x}+ \mathbf{f}_{b}$. By querying these points, the corresponding color, density, and forward/backward flow can be predicted by the same MLP:
\begin{equation}
\begin{split}
\begin{aligned}
    (\mathbf{c}_{d}^{t+1},\sigma_{d}^{t+1},\mathbf{f}_{ff},\mathbf{f}_{fb})=\texttt{MLP}_d(\mathbf{x}+ \mathbf{f}_{f},\mathbf{d},t+1), \\
    (\mathbf{c}_{d}^{t-1},\sigma_{d}^{t-1},\mathbf{f}_{bf},\mathbf{f}_{bb})=\texttt{MLP}_d(\mathbf{x}+ \mathbf{f}_{b},\mathbf{d},t-1).
\end{aligned}
\end{split}
\end{equation}
The rendering color $\mathbf{\widehat{C}}^{\overline{t}}_d$ of the ray $\mathbf{r}$ in the consequent three timestamps $\overline{t} \in \left\{t-1,t,t+1\right\}$  can be computed with the volume rendering integral. NSFF can be trained with minimizing 
\begin{equation}
    \mathcal{L}_{dynamic} =\sum_{\overline{t} \in \left\{t-1,t,t+1\right\}} \sum_{\mathbf{r}\in \mathcal{R} } \| \mathbf{\widehat{C}}_d^{\overline{t}}(\mathbf{r})- \mathbf{C}(\mathbf{r}) \|_{2}^{2}.
\end{equation}
At the same time, the scene flow $\mathbf{f}_f$ and $\mathbf{f}_b$ are supervised with the pre-processed 2D optical flow by an off-the-shelf  method during training. Specifically, 
 each 3D position corresponding to 2D pixel at time $t$ is warped to the reference frame via 
 $\mathbf{f}_f$ and $\mathbf{f}_b$ 
 and then projected 
 onto the reference camera to generate 2D optical flow, which is enforced to approach the pre-processed 2D optical flow by a pre-trained network.
Also, the expected termination depth along each ray and the predicted dynamic probability $p$ are supervised with
the pre-processed depth and mask by off-the-shelf methods, respectively.

In addition, the NSFF model is also regularized by the following constraints.  
$\|\mathbf{f}_f\|_1$ and $\|\mathbf{f}_b\|_1$ are minimized to promote the predicted scene flow to be temporally smooth and slow, and  
$\|\mathbf{f}_f+\mathbf{f}_{fb}\|_2$ and $\|\mathbf{f}_b+\mathbf{f}_{bf}\|_2$ are minimized to promote the scene flow to be consistent. 
The depth maps from the static NeRF are used to restrict those from NSFF. To ensure that only a few samples representing the dynamic objects dominate the rendering of the NSFF, the entropy of the rendering weights $T_d$ along each ray is minimized.

Finally, the static NeRF and NSFF are fused by leading the predicted dynamic probability value $p$ into the rendering equation to produce a composite color:
\begin{equation}
\begin{split}
    \mathbf{\widehat{C}}_{f}(r) = \int_{s_{n}}^{s_{f}}T_{f}(s) \big((1-p)\sigma_s(\mathbf{r}(s))\mathbf{c}_s(\mathbf{r}(s),\mathbf{d})\\
    +p\sigma_d(\mathbf{r}(s))\mathbf{c}_d(\mathbf{r}(s),\mathbf{d})\big)dt, 
\end{split}
\end{equation}
where 
\begin{equation}
T_f=\texttt{Exp}(-{\int_{s_{n}}^{s}}\big((1-p)\sigma_s(\mathbf{r}(q))\big)\big(p \sigma_d (\mathbf{r}(q))\big)dq),
\end{equation}
and $s_n$, and $s_f$ are the near and far bounds of the scene, respectively. Also, $\widehat{\mathbf{C}}_f$ is supervised with the ground-truth, i.e.,  
\begin{equation}
    \mathcal{L}_{full} = \sum_{\mathbf{r}\in \mathcal{R} }  \| \mathbf{\widehat{C}}_f(\mathbf{r})- \mathbf{C}(\mathbf{r}) \|_{2}^{2}.
\end{equation}
\section{Proposed Method}
\label{sec:method}

\label{decoupling}

\begin{figure}[t]
\centering
\includegraphics[width=0.47\textwidth]{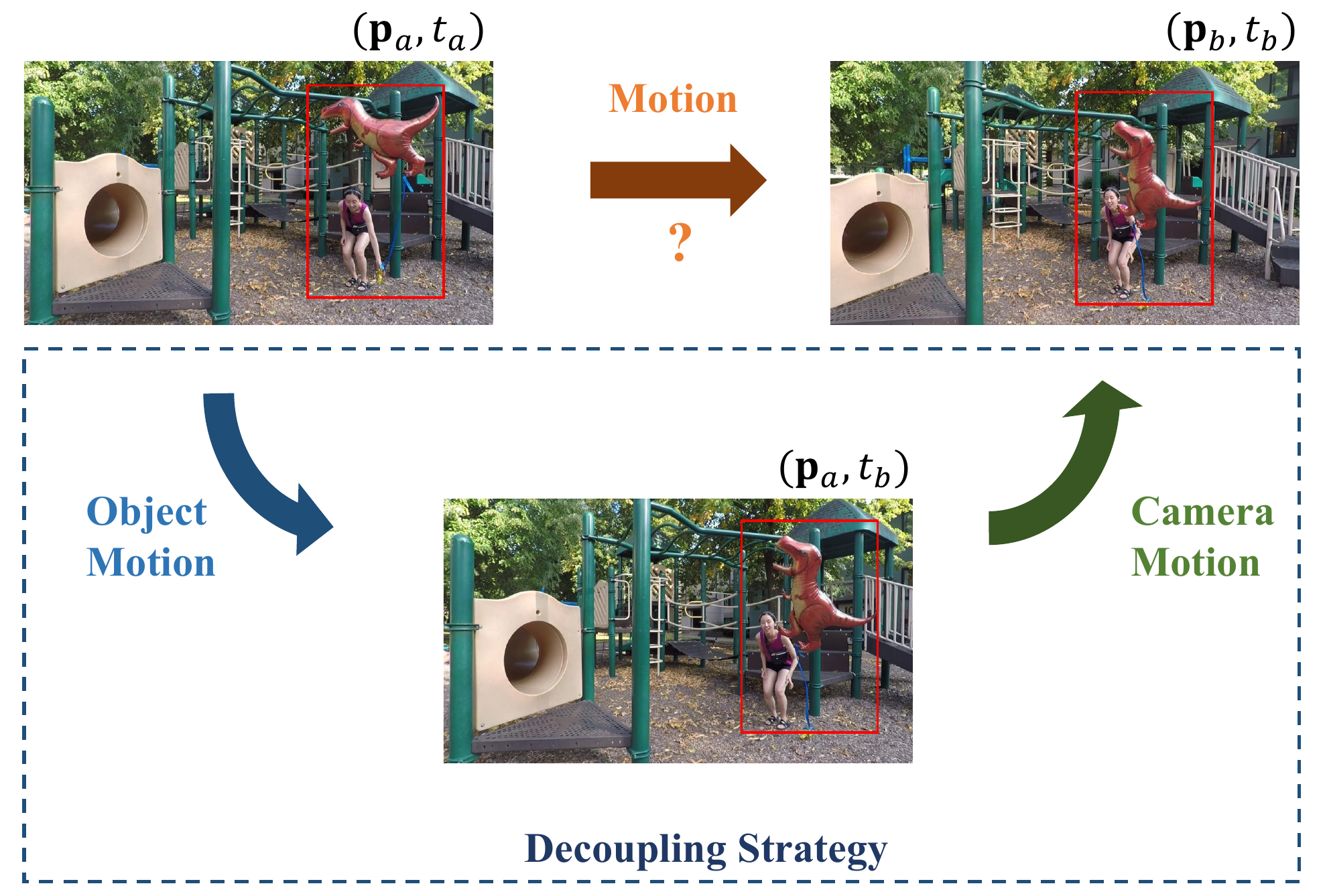}\vspace{-0.4cm}
\caption{ 
Given two successive frames of a dynamic monocular video with their camera poses and timestamps being  $(\mathbf{p}_a,~t_a)$ and $(\mathbf{p}_b,~t_b)$. Directly modeling the motion of \textit{the dynamic objects} (highlighted in \textcolor{red}{red frames}) between them can be challenging. 
Here, we propose to decouple the motion between them into two types:
(\textbf{1}) the motion caused by the object movement; and (\textbf{2}) the motion caused by the camera movement. }
\label{figure: decoupling}
\end{figure}

\noindent \textbf{Motivation}. As aforementioned, to tackle the challenge of modeling the \textit{dynamic objects} of a scene, existing dynamic view synthesis methods  
adopt pre-processed 2D optical flow and depth maps by off-the-shelf methods to supervise the 
NSFF explicitly. However, such a manner suffers from the following drawbacks. 
The pre-processed maps inevitably contain inaccurate regions, which would be harmful to the network during training, and thus compromise the quality of synthesized views, as demonstrated in Fig. \ref{fig:drawbacks}. Moreover, 
 the ambiguity exists  when using the pre-processed 2D information to supervise the 3D geometry of the scene, as the same 2D pixels can result from various density and radiance distributions on the ray due to the rendering integral, leading to an unlimited number of possible explanations. Finally,  obtaining accurate pre-processed maps may be computationally expensive and time-consuming, making it inconvenient.\\

\noindent\textbf{Our Solution}. 
Unlike existing methods, we tackle this challenging problem in an \textit{unsupervised} manner.  
Generally, as shown in Fig.~\ref{figure: decoupling}, we propose to decouple the motion of dynamic objects between two successive frames of a monocular video, 
such that the transition/flow from $\mathbf{I}_a$ at $(\mathbf{p}_a,t_a)$ to $\mathbf{I}_b$ at $(\mathbf{p}_b,t_b)$ can be modeled by first transforming the timestamp only with object motion involved, 
and then transforming the pose coordinate only with camera motion involved. 
Such a decoupling process is natural for monocular videos and can produce a fine-grained formulation of the motion of dynamic objects  to alleviate the network's difficulty in learning their complex 3D motion from extremely limited 2D observations. 
In accordance with our decoupling modeling approach, we propose two \textit{unsupervised} regularization terms to enforce the network to learn under this paradigm. These terms are the surface consistency constraint (Sec. \ref{subsec:constraint1}), which addresses temporal changes, and the patch-based multi-view constraint (Sec. \ref{subsec:constraint2}), which addresses camera changes.
Note that our method is built upon Gao \textit{et al.} \cite{gao2021dynamic} without using pre-processed optical flow and depth as supervision.   In the following, we will detail the two proposed constraints.

\begin{figure}[t]
\centering
\includegraphics[width=0.45\textwidth]{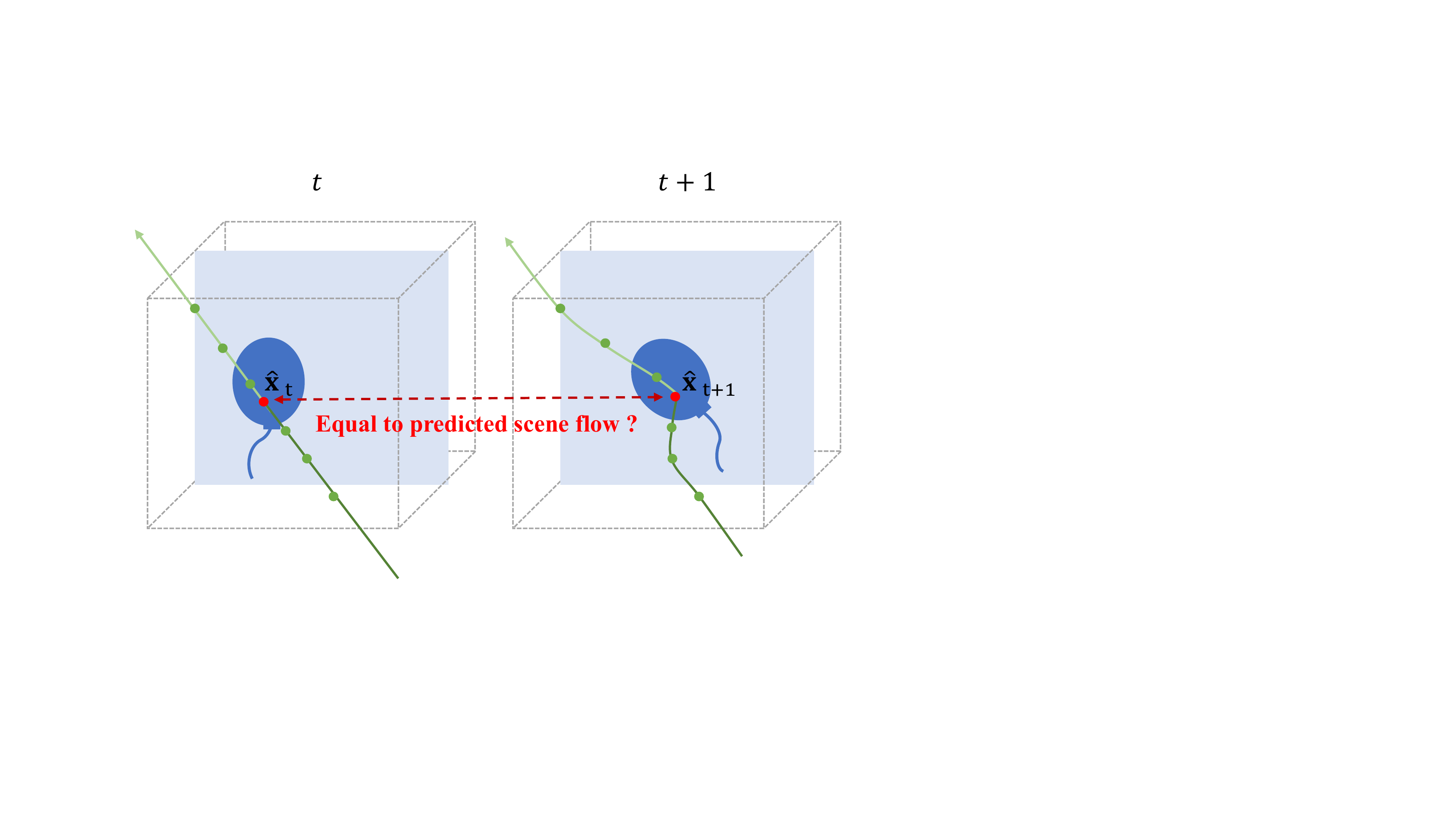}
\caption{
Illustration of the surface consistency constraint. Under the NeRF scheme,  at timestamp $t$, we sample points (\textcolor{green}{green} points) on a ray and predict their density values to compute the weights they contribute to the rendering. The intersection point $\hat{\mathbf{x}}_t$ 
between the surface and the ray can be calculated by a weighted averaging of the coordinates of the sampled points. With NSFF, we can predict a 3D scene flow for each of the sampled points and warp them to the next timestamp $t+1$. Also, we can calculate the intersection point $\hat{\mathbf{x}}_{t+1}$ between the warped ray and the surface 
by a weighted averaging of the coordinates of the warped points. Our surface consistency constraint requires the difference between $\hat{\mathbf{x}}_{t}$ and $\hat{\mathbf{x}}_{t+1}$ 
should equal the scene flow predicted for the  intersection point at $t$ by NSFF.
}
\label{figure: surface}
\end{figure}

\subsection{Surface Consistency Constraint}
\label{subsec:constraint1}

Assuming that the 3D geometric surface  of a moving object is dynamically consistent throughout the video, we propose the surface consistency constraint to  regularize the object motion among different timestamps,
 That is, the surface in the current timestamp should be correctly mapped to scene surfaces in its neighboring timestamps using the NSFF. 
 To be more specific, we ensure that the predicted flow for a point on the scene surface at time $t$ accurately maps that point to the surface at time $t+1$.

Specifically, as illustrated in Fig.~\ref{figure: surface}, for the ray $\mathbf{r}$ at time $t$, we sample $K$ points $\left\{\mathbf{x}_i\right\}_{i=1}^K$ for NeRF optimization and predict their corresponding density value, denoted as $\{\sigma_{d,i}^t\}_{i=1}^K$. For each point, we calculate the weight indicating its contribution to the rendering as
\begin{equation}
    w_{d,i}^{t} = T_d \sigma_{d,i}^{t},
\end{equation}
where 
\begin{equation}
    T_{d,i}=\texttt{Exp}(-{\int_{s_{n}}^{s_i}} \sigma_d^t (\mathbf{r}(q))dq),
\end{equation}
where $s_i$ is the distance between $\mathbf{x}_i$ and $\mathbf{o}$. 
Also, for $\mathbf{r}$ at time $t$, we predict its intersection point with the scene surface, denoted as $\widehat{\mathbf{x}}_t$, by a weighted average of the coordinates of the sampled points over $\mathbf{r}$: 
\begin{equation}
    \widehat{\mathbf{x}}_t=\sum_{i=1}^K w_{d,i}^{t}\cdot \mathbf{x}_i.
\end{equation}
With the scene flow $\mathbf{f}_{f}$ by NSFF (i.e., Eq.~\eqref{dynamicnerf}), we can warp 
ray $\mathbf{r}$ 
to the next timestamp $t+1$: 
\begin{equation}
    \widetilde{\mathbf{x}}_i = \mathbf{x}_i+\mathbf{f}_{f}(\mathbf{x}_i), 
\end{equation}
where $\mathbf{f}_{f}(\mathbf{x}_i)$ refers to the scene flow for $\mathbf{x}_i$, and $\widetilde{\mathbf{x}}_i$ is the warped point of $\mathbf{x}_i$. \textit{Note that the scene flow} $\mathbf{f}_{f}$ \textit{is only an output of NSFF and NOT supervised during training in our framework}. 
Similar to the above process, for $\widetilde{\mathbf{x}}_i$, we can also predict the corresponding weight $w_{d,i}^{t+1}$ and the intersection point of the warped ray with the scene surface at $t+1$, denoted as $\widehat{\mathbf{x}}_{t+1}$: 
\begin{equation}
    \widehat{\mathbf{x}}_{t+1}=\sum_{i=1}^K w_{d,i}^{t+1}\cdot \widetilde{\mathbf{x}}_i. 
\end{equation}
Based on the fact that the scene flow-induced surface should be consistent with the predicted surface, we propose a point-to-point loss by enforcing the warped intersection point of $\widehat{\mathbf{x}}_t$ by the flow, denoted as  $\widehat{\mathbf{x}}_t+\mathbf{f}_f(\widehat{\mathbf{x}}_t)$, to coincide with 
$\widehat{\mathbf{x}}_{t+1}$, i.e., 
\begin{equation}
    \mathcal{L}_{surface} = \sum_{\mathbf{r}\in \mathcal{R} } \left\| (\widehat{\mathbf{x}}_t+\mathbf{f}_f(\widehat{\mathbf{x}}_t)) - \widehat{\mathbf{x}}_{t+1} \right\|_{1}, 
\end{equation}
where $\left\|\cdot\right\|_1$ is the $\ell_1$-norm of a vector.
We also apply this loss to the previous timestamp $t-1$. 

\begin{figure}[t]
\centering
\includegraphics[width=0.45\textwidth]{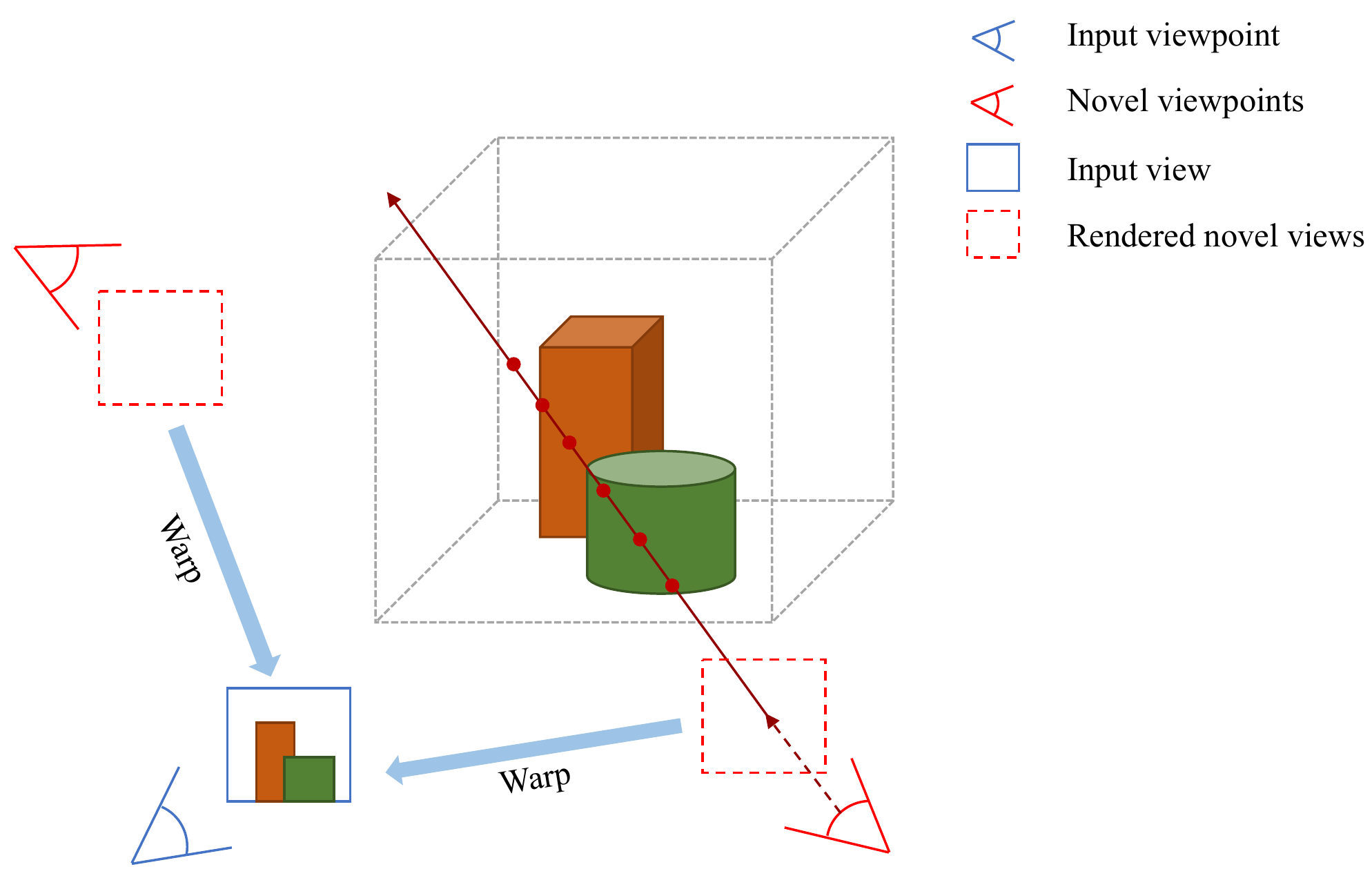}
\caption{
Illustration of the patch-based multi-view constraint. 
Under a certain timestamp, we enforce the multi-view consistency between the single input view and unknown novel viewpoints. We can synthesize a well-quality RGB-D image at the input viewpoint (\textcolor{blue}{blue} viewpoint) owing to the capacity of NeRF.
We then randomly sample novel viewpoints (red viewpoints) and render novel views in patch-wise for \textit{efficiency}. The rendered novel view is warped to the input viewpoint using the rendered depth map for the input viewpoint. We compute the photometric reconstruction loss between the warped and rendered \textit{patches} of the input view for efficiency.
}
\label{figure: patch}
\end{figure}

\subsection{Patch-based Multi-view Constraint}
\label{subsec:constraint2}

Here we propose the multi-view constraint to ensure consistency between the single input view and synthesized novel views at the same timestamp, enhancing the ability of the network to handle various camera movements. Moreover, we implement this constraint patch-wise for efficiency.

Taking time $t_m$ ($m=1,~2,~\cdots,~N$) as an example. 
Specifically, as shown in Fig.~\ref{figure: patch}, at $t_m$, we have a single input view $\mathbf{I}_m$ associated with its camera pose $\mathbf{p}_m$ available. Owing to the powerful capability of NeRF, we can synthesize a well-quality patch $\mathbf{I}_s^p$ and derive a reasonable depth map $\mathbf{D}_ s^p$ for this known viewpoint.
We assume that this RGB-D map accurately represents the appearance and geometry of the scene and can thus serve as supervision for the synthesized novel views. To train a robust model, we randomly select a viewpoint $\mathbf{p}_r$ as the object of supervision and render a new patch $\mathbf{I}_r^p$. With $\mathbf{D}_s^p$, we inversely warp $\mathbf{I}_r^p$ to the viewpoint $\mathbf{p}_m$, written as 
\begin{equation}
    \widetilde{\mathbf{I}}_s^p= \mathcal{W}(\mathbf{I}_r^p,~\mathbf{D}_s^p,~\mathbf{p}_r,~\mathbf{p}_m),
\end{equation}
where $\mathcal{W}(\cdot)$ denotes the inverse warping operation.
Finally, we formulate the patch-based multi-view constraint as 
\begin{equation}
    \mathcal{L}_{patch}= \| \mathbf{M_w}\odot(\mathbf{I}_s^p - \widetilde{\mathbf{I}}_s^p )\|_1,
\end{equation}
where $\mathbf{M_w}$ is the induced mask by the inverse warping operation, and $\odot$ denotes the Hadamard product.


\begin{table}[t]
\centering
\begin{center}
\caption{ Pre-processed optical flow maps and depth maps supervision used in different methods. ``$\checkmark$"  (resp. ``$\times$") represents the corresponding term is used (resp. unused).}
\vspace{-0.4cm}
\label{table:supervision}
 \resizebox{0.5\textwidth}{!}{
\begin{tabular}{c|c|c}
\toprule[1.5pt]
&Optical flow maps &Depth maps  
\\\hline
Yoon \textit{et al.} \cite{yoon2020novel}&$\checkmark$&$\checkmark$
\\
Tretschk \textit{et al.} \cite{tretschk2021non}&$\times$&$\times$
\\
Li \textit{et al.} \cite{li2021neural}&$\checkmark$&$\checkmark$
\\
Gao \textit{et al.} \cite{gao2021dynamic}&$\checkmark$&$\checkmark$
\\
Yang \textit{et al.} \cite{yang2023deformable}&$\times$&$\times$
\\
Ours &$\times$ &$\times$\\
\bottomrule[1.5pt]
\end{tabular}
}
\end{center} 
\end{table}

\begin{table*}[t]
\centering
\begin{center}
\caption{Quantitative comparison of different methods on the Dynamic Scene Dataset. $\uparrow$ (resp. $\downarrow$) means the larger (resp. smaller), the better. The best and second best results are highlighted in \textbf{bold} and \underline{underline}, respectively. 
} \vspace{-0.3cm}
\label{tabel:results}
\resizebox{0.9\textwidth}{!}{
\begin{tabular}{l|c|c|c|c}
\toprule[1.2pt]
  PSNR$\uparrow$ / SSIM$\uparrow$ / LPIPS$\downarrow$  & Balloon1 & Balloon2   & Jumping & Playground \\\hline

NeRF\cite{mildenhall2020nerf} & 19.82 / 0.699 / 0.287& 24.37 / 0.815 / 0.180&20.99 / 0.687 / 0.418&21.07 / 0.768 / 0.240
\\
NeRF-t  & 18.54 / 0.443 / 0.419&20.69 / 0.570 / 0.349&18.04 / 0.465 / 0.535&14.68 / 0.230 / 0.534
\\
Yoon \textit{et al.} \cite{yoon2020novel} &18.74 / 0.606 / 0.248&19.88 / 0.416 / 0.288&20.15 / 0.616 / 0.250&15.08 / 0.249 / 0.309
\\
Tretschk \textit{et al.} \cite{tretschk2021non} & 17.39 / 0.363 / 0.496&22.41 / 0.708 / 0.319&20.09 / 0.630 / 0.416&15.06 / 0.240 / 0.458
\\
Li \textit{et al.} \cite{li2021neural} & 21.96 / \underline{0.702} / 0.288&24.27 / 0.740 / 0.288&\underline{24.65} / 0.813 / 0.227&21.22 / 0.717 / 0.268
\\
Gao \textit{et al.} \cite{gao2021dynamic} & \underline{22.36} / \textbf{0.773} / \textbf{0.173}&\underline{27.06} / \underline{0.864} / \textbf{0.117}&\textbf{24.68} / \textbf{0.843} / \textbf{0.151}&\underline{24.15} / \underline{0.858} / \underline{0.148}
\\
Yang \textit{et al.} \cite{yang2023deformable} & 20.44 / 0.634 / 0.234 & 24.67 / 0.790 / 0.172 &21.49 / 0.691 / 0.311&21.58 / 0.778 / 0.153
\\
Ours & \textbf{22.49} / \textbf{0.773} / \underline{0.186}&\textbf{27.34} / \textbf{0.870} / \underline{0.120}&24.09 / \underline{0.823} / \underline{0.162}&\textbf{24.62} / \textbf{0.864} / \textbf{0.147}
\\ \hline
\hline
  PSNR$\uparrow$ / SSIM$\uparrow$ / LPIPS$\downarrow$  &  Skating & Truck & Umbrella & Average \\\hline

NeRF \cite{mildenhall2020nerf} &23.67 / 0.781 / 0.418& 22.73 / 0.747 / 0.334&21.29 / 0.546 / 0.457 & 21.99 / 0.720 / 0.333
\\
NeRF-t  &20.32 / 0.504 / 0.558& 18.33 / 0.4380 / 0.524&17.69 / 0.267 / 0.667& 18.33 / 0.417 / 0.512
\\
Yoon \textit{et al.} \cite{yoon2020novel} &21.75 / 0.561 / 0.262&21.53 / 0.623 / \underline{0.172}&20.35 / 0.543 / 0.243& 19.64 / 0.517 / 0.253
\\
Tretschk \textit{et al.} \cite{tretschk2021non} &23.95 / 0.732 / 0.343& 19.33 / 0.462 / 0.546&19.63 / 0.380 / 0.480& 19.69 / 0.502 / 0.437
\\
Li \textit{et al.} \cite{li2021neural} &29.29 / 0.888 / 0.190& 25.96 / 0.774 / 0.249& 22.97 / 0.659 / 0.319& 24.33 / \underline{0.756} / 0.261
\\
Gao \textit{et al.} \cite{gao2021dynamic} &\underline{32.66} / \underline{0.951} / \textbf{0.080}& \underline{28.56} / \textbf{0.872} / \textbf{0.149}&\underline{23.26 }/ \underline{0.720} / \underline{0.193}& \underline{26.10} / \textbf{0.840} / \textbf{0.144}
\\
Yang \textit{et al.} \cite{yang2023deformable} &24.09 / 0.742 / 0.295&25.09 / 0.7478 / 0.234 &22.10 / 0.625 / 0.231& 22.78 / 0.715 / 0.233\\
Ours &\textbf{33.66} / \textbf{0.955} / \underline{0.084}& \textbf{28.63} / \underline{0.871} / \underline{0.172}&\textbf{23.71} / \textbf{0.726} / \textbf{0.186}& \textbf{26.36} / \textbf{0.840} / \underline{0.151}
\\ \bottomrule[1.2pt]

\hline
\end{tabular}
}
\end{center} 
\end{table*}

\begin{table*}[h]
\centering
\begin{center}
\caption{Quantitative comparison of different methods on the \textbf{dynamic part} of a scene. $\uparrow$ (resp. $\downarrow$) means the larger (resp. smaller), the better. The best and second best results are highlighted in \textbf{bold} and \underline{underline}, respectively. 
} \vspace{-0.3cm}
\label{tabel:results_d}
\resizebox{0.9\textwidth}{!}{
\begin{tabular}{l|c|c|c|c}
\toprule[1.2pt]
  PSNR$\uparrow$ / SSIM$\uparrow$ / LPIPS$\downarrow$  & Balloon1 & Balloon2   & Jumping & Playground \\\hline

NeRF\cite{mildenhall2020nerf} & 17.06 / 0.562 / 0.395& 19.74 / 0.631 / 0.273&16.95 / 0.471 / 0.480&16.62 / 0.618 / 0.337
\\
NeRF-t  & 17.01 / 0.375 / 0.447&17.47 / 0.316 / 0.469&14.08 / 0.248 / 0.578&13.99 / 0.147 / 0.571
\\
Yoon \textit{et al.} \cite{yoon2020novel} &17.46 / 0.505 / 0.318&18.59 / 0.332 / 0.272&17.17 / 0.466 / 0.289&13.86 / 0.188 / 0.402
\\
Tretschk \textit{et al.} \cite{tretschk2021non} & 16.69 / 0.341 / 0.531&18.69 / 0.531 / 0.425& 15.88 / 0.357 / 0.514&13.83 / 0.136 / 0.573
\\
Li \textit{et al.} \cite{li2021neural} &\underline{19.73} / 0.609 / 0.334 & 21.57 / 0.590 / 0.351& \textbf{20.61} / \underline{0.668 }/ 0.286&\underline{20.09} / 0.697 / 0.275
\\
Gao \textit{et al.} \cite{gao2021dynamic} &19.49 / \underline{0.647} / \textbf{0.251} &\underline{22.93 }/ \underline{0.738} / \textbf{0.157}&\underline{19.95 }/ \textbf{0.669} / \textbf{0.237}&19.76 / \underline{0.726} / \underline{0.221}
\\

Yang \textit{et al.} \cite{yang2023deformable} &  19.11 / 0.548 / 0.322 & 22.36 / 0.691 / 0.227 & 18.21 / 0.512 / 0.391& 18.75 / 0.652 / 0.260
\\
Ours &\textbf{19.80} / \textbf{0.653 }/\underline{ 0.265 }&\textbf{23.45 }/ \textbf{0.763} / \underline{0.175}& 19.46 / 0.629 / \underline{0.262}& \textbf{20.63} / \textbf{0.753} / \textbf{0.195}
\\ \hline
\hline
  PSNR$\uparrow$ / SSIM$\uparrow$ / LPIPS$\downarrow$  &  Skating & Truck & Umbrella & Average \\\hline

NeRF \cite{mildenhall2020nerf} &14.44 / 0.406 / 0.563& 15.40 / 0.370 / 0.471&13.96 / 0.290 / 0.624& 16.31 / 0.478 / 0.449
\\
NeRF-t  &14.27 / 0.284 / 0.648& 15.58 / 0.278 / 0.481&14.29 / 0.188 / 0.720& 15.24 / 0.262 / 0.559
\\
Yoon \textit{et al.} \cite{yoon2020novel} &15.00 / 0.251 / 0.358&19.32 / 0.523 / 0.207&14.50 / 0.326 / 0.425& 16.56 / 0.370 / 0.324
\\
Tretschk \textit{et al.} \cite{tretschk2021non} &14.37 / 0.324 / 0.561& 16.95 / 0.304 / 0.540&14.04 / 0.208 / 0.647& 15.78 / 0.315 / 0.542
\\
Li \textit{et al.} \cite{li2021neural} &21.65 / 0.651 / 0.271&\underline{23.38} / 0.689 / 0.213 & \underline{15.90 }/ 0.401 / 0.463& \underline{20.42} / 0.615 / 0.313
\\
Gao \textit{et al.} \cite{gao2021dynamic} &\underline{21.68 }/\underline{ 0.680} /\textbf{ 0.201}&23.21 /\underline{ 0.706} / \textbf{0.121}&15.79 /\underline{ 0.428} / \underline{0.361}&  20.40 / \underline{0.656} / \textbf{0.221}
\\
Yang \textit{et al.} \cite{yang2023deformable} & 15.80 / 0.393 / 0.494 & 22.59 / 0.590 / 0.392& 16.17 / 0.376 / 0.468&  19.00 / 0.538 / 0.365\\
Ours &\textbf{23.31} / \textbf{0.771 }/\underline{ 0.208}&\textbf{23.65 }/\textbf{ 0.714} / \underline{0.190} &\textbf{16.52 }/\textbf{ 0.470 }/ \textbf{0.320}& \textbf{20.97} / \textbf{0.679} / \underline{0.231}
\\ \bottomrule[1.2pt]

\hline
\end{tabular}
}
\end{center} 
\end{table*}

\begin{figure*}[t]
\centering
\includegraphics[width=0.8\textwidth]{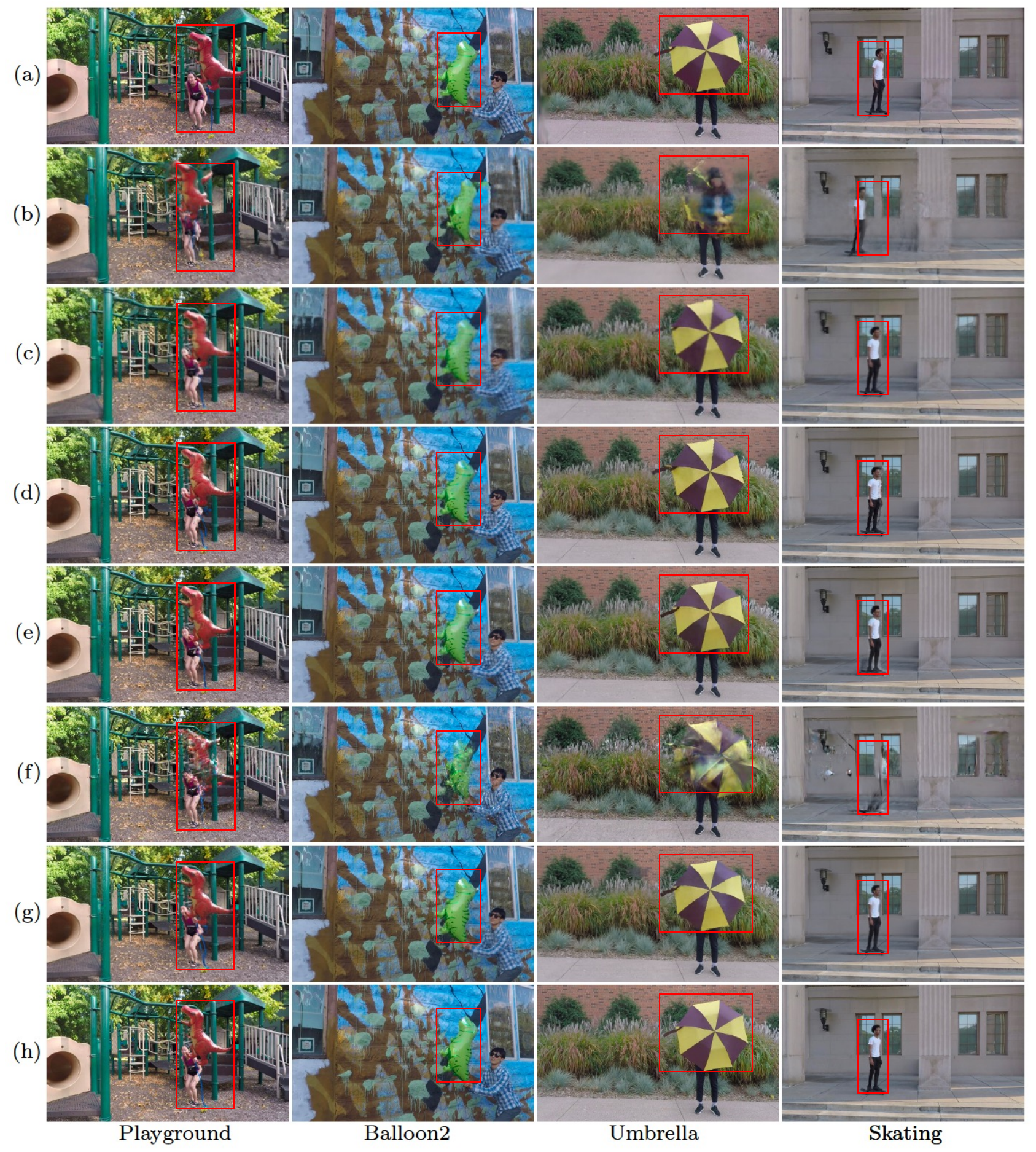} \vspace{-0.4cm}
\caption{Visual comparison of different methods on Dynamic Scene Dataset. (a) Yoon \textit{et al.} \cite{yoon2020novel}, (b) Tretschk \textit{et al.} \cite{tretschk2021non}, (c) Li \textit{et al.} \cite{li2021neural}, (d) Gao \textit{et al.} \cite{gao2021dynamic}, (e) Yang \textit{et al.} \cite{yang2023deformable}, (f) Ours, (g) Ground Truth. Dynamic regions are highlighted in the red rectangular.}
\label{figure: result}
\end{figure*}

\textbf{Remark}.  The concept of utilizing multi-view consistency to enhance NeRF in handling sparse views is similar between our method and RegNeRF \cite{Niemeyer2021Regnerf}. However, the specific idea of the patch-based multi-view constraint in our method differs significantly from that of RegNeRF \cite{Niemeyer2021Regnerf}
In our approach, we focus on using a single input view to supervise other viewpoints within a specific timestamp. We achieve this by warping unknown viewpoints, represented as patches, to align with the input view and then applying RGB consistency loss for supervision. On the other hand, RegNeRF samples unobserved viewpoints in 3D space and directly renders a patch from the selected viewpoint. They utilize depth smoothness loss and RGB log-likelihood loss for their training process. While both methods share the idea of multi-view consistency, our patch-based approach involves a warping process that distinguishes it from RegNeRF's methodology, which primarily relies on depth smoothness and RGB log-likelihoods without warping.

\section{Experiments}
\label{sec: experiments}
\subsection{Experiment Settings}

\noindent\textbf{Dataset.}
We conducted extensive experiments on the NVIDIA Dynamic Scene Dataset \cite{yoon2020novel} and the Neural 3D Video Synthesis Dataset \cite{li2022neural}. For each scene in the NVIDIA Dynamic Scene Dataset, the dataset provides 12 sequences captured by 12 fixed cameras. The scene contains a static background and a moving object in the foreground. We followed the setting of training/testing data and pre-processing protocol of Gao \textit{et al.} \cite{gao2021dynamic}.  The input video is obtained by selecting the $i$-th view at time $t_i$. In this way, each input view has a different viewpoint and timestamp from others. The evaluation set is determined by fixing the camera to the first view and changing timestamps. The Neural 3D
Video Synthesis Dataset, which features casual videos that depict a man engaged in activities such
as cooking or preparing coffee. This dataset encompasses 6 sequences for each scene, recorded
using multiple stationary cameras. We adhered to the training/testing data configuration and the preprocessing
guidelines as specified in the Dynamic Scene Dataset.\\ 

\noindent\textbf{Implementation details.}
Our approach follows NeRF to use positional encoding for input coordinates. For NSFF, we add an extra positional encoding of time. We sample 64 points along each ray for both training and testing. We first train the static NeRF and then fix its network parameters when training the NSFF. Training a full model takes approximately one day per scene using one NVIDIA GeForce RTX 3090 GPU, and rendering a single frame takes roughly 3 seconds.

\subsection{Comparison with State-of-the-Art Methods}
We compared our method with six methods: NeRF \cite{mildenhall2020nerf}, NeRF-t (a straightforward extension of NeRF by incorporating the time dimension into the input coordinate), an image-based rendering method  Yoon \textit{et al.} \cite{yoon2020novel}, a deformable NeRF-based method Tretschk \textit{et al.} \cite{tretschk2021non}, two NSFF-based methods Li \textit{et al.} \cite{li2021neural} and Gao \textit{et al.} \cite{gao2021dynamic}, and a 3D-GS based method Yang \textit{et al.} \cite{yang2023deformable}. 
We also summarize whether they use pre-processed optical flow or depth as supervision in Table \ref{table:supervision}.\\

\begin{table*}[t]
\centering
\begin{center}
\caption{Quantitative comparison of different methods on the Neural 3D Video Synthesis Dataset. $\uparrow$ (resp. $\downarrow$) means the larger (resp. smaller), the better. The best and second best results are highlighted in \textbf{bold} and \underline{underline}, respectively. } \vspace{-0.3cm}
\label{tabel:results_3dv}
\resizebox{0.9\textwidth}{!}{
\begin{tabular}{l|c|c|c|c}
\toprule[1.2pt]
  PSNR$\uparrow$ / SSIM$\uparrow$ / LPIPS$\downarrow$  & Li \textit{et al.} \cite{li2021neural} & Gao \textit{et al.} \cite{gao2021dynamic}&Yang \textit{et al.} \cite{yang2023deformable}&Ours \\\hline

coffee martini &  20.51 / 0.672 / 0.378  & \underline{28.20} / \underline{0.909} / \underline{0.146}&23.73 / 0.842 / 0.177& \textbf{29.94} / \textbf{0.930} / \textbf{0.117}\\
cut roosted beef   &23.51 / 0.779 / 0.257&\underline{27.93} / \underline{0.900} / \underline{0.132}&23.44 / 0.845 / 0.179&\textbf{28.90} / \textbf{0.910} / \textbf{0.124}\\
cook spinach &26.68 / 0.843 / 0.264&\underline{27.15} / \underline{0.897} / \underline{0.141}&26.67 / 0.872 / 0.145&\textbf{29.16} / \textbf{0.917} / \textbf{0.119}\\
flame steak &23.00 / 0.766 / 0.286 &\underline{26.46} / \underline{0.903} / \underline{0.141} &25.97 / 0.836 / 0.150&\textbf{28.39} / \textbf{0.922} / \textbf{0.121}\\
sear steak & 25.05 / 0.795 / 0.276&\underline{28.23 }/ \underline{0.914} / \underline{0.126}&22.67 / 0.792 / 0.212& \textbf{29.42} / \textbf{0.924} / \textbf{0.117}\\
flame salmon  &19.73 / 0.705 / 0.352& \underline{24.53} / \underline{0.885 }/ \underline{0.150} &22.30 / 0.810 / 0.189 &  \textbf{25.90} / \textbf{0.898} / \textbf{0.137}\\ \hline
Average &23.08 / 0.760 / 0.302&\underline{27.08} / \underline{0.901} / \underline{0.139}&24.13 / 0.833 / 0.175 &\textbf{28.62} / \textbf{0.917} / \textbf{0.123}
\\ \bottomrule[1.2pt]
\hline
\end{tabular}
}
\end{center} 
\end{table*}

\begin{table*}[t]
\centering
\begin{center}
\caption{Quantitative comparison of different methods on the \textbf{dynamic part} of a scene on the Neural 3D Video Synthesis Dataset. $\uparrow$ (resp. $\downarrow$) means the larger (resp. smaller), the better. The best and second best results are highlighted in \textbf{bold} and \underline{underline}, respectively. }\vspace{-0.3cm}
\label{tabel:results_3dv_dy}
\resizebox{0.9\textwidth}{!}{
\begin{tabular}{l|c|c|c|c}
\toprule[1.2pt]
PSNR$\uparrow$ / SSIM$\uparrow$ / LPIPS$\downarrow$  & Li \textit{et al.} \cite{li2021neural} & Gao \textit{et al.} \cite{gao2021dynamic}&Yang \textit{et al.} \cite{yang2023deformable}&Ours \\\hline
coffee martini &  18.23 / 0.506 / 0.438&\underline{21.51} / \underline{0.689} / \underline{0.290}& 17.48 / 0.591 / 0.344& \textbf{23.61} / \textbf{0.791} / \textbf{0.207}\\
cut roosted beef  & 19.46 / 0.521 / 0.409 &\underline{21.94} / \underline{0.667} / \underline{0.279}& 18.31 / 0.597 / 0.367&\textbf{23.31} /\textbf{ 0.717} / \textbf{0.258}\\
cook spinach & 21.46 / 0.644 / 0.386&20.76 / 0.648 / 0.303&\underline{22.02} /\underline{ 0.685} /\underline{ 0.281}&\textbf{23.53 }/ \textbf{0.745} / \textbf{0.235}\\
flame steak &18.50 / 0.540 / 0.422&19.57 / 0.638 / 0.333 &\textbf{22.30} / \underline{0.726} /\underline{ 0.261}&\underline{21.82} / \textbf{0.730} / \textbf{0.246}\\
sear steak & 21.27 / 0.597 / 0.400&\underline{21.90} / \underline{0.694} / \underline{0.292}&17.10 / 0.506 / 0.472&\textbf{23.55} / \textbf{0.761} / \textbf{0.228}\\
flame salmon & 15.24 / 0.399 / 0.508&\underline{17.56} / \underline{0.569} / \underline{0.365} & 16.70 / 0.502 / 0.443& \textbf{19.09} / \textbf{0.638} / \textbf{0.309}\\ \hline
Average & 19.03 / 0.535 / 0.427 & \underline{20.54} /\underline{ 0.651} / \underline{0.310} & 18.99 / 0.601 / 0.361& \textbf{22.49} / \textbf{0.730} / \textbf{0.247}\\
\bottomrule[1.2pt]
\hline
\end{tabular}
}
\end{center} 
\end{table*}

\begin{figure*}[t]
\centering
\includegraphics[width=0.8\textwidth]{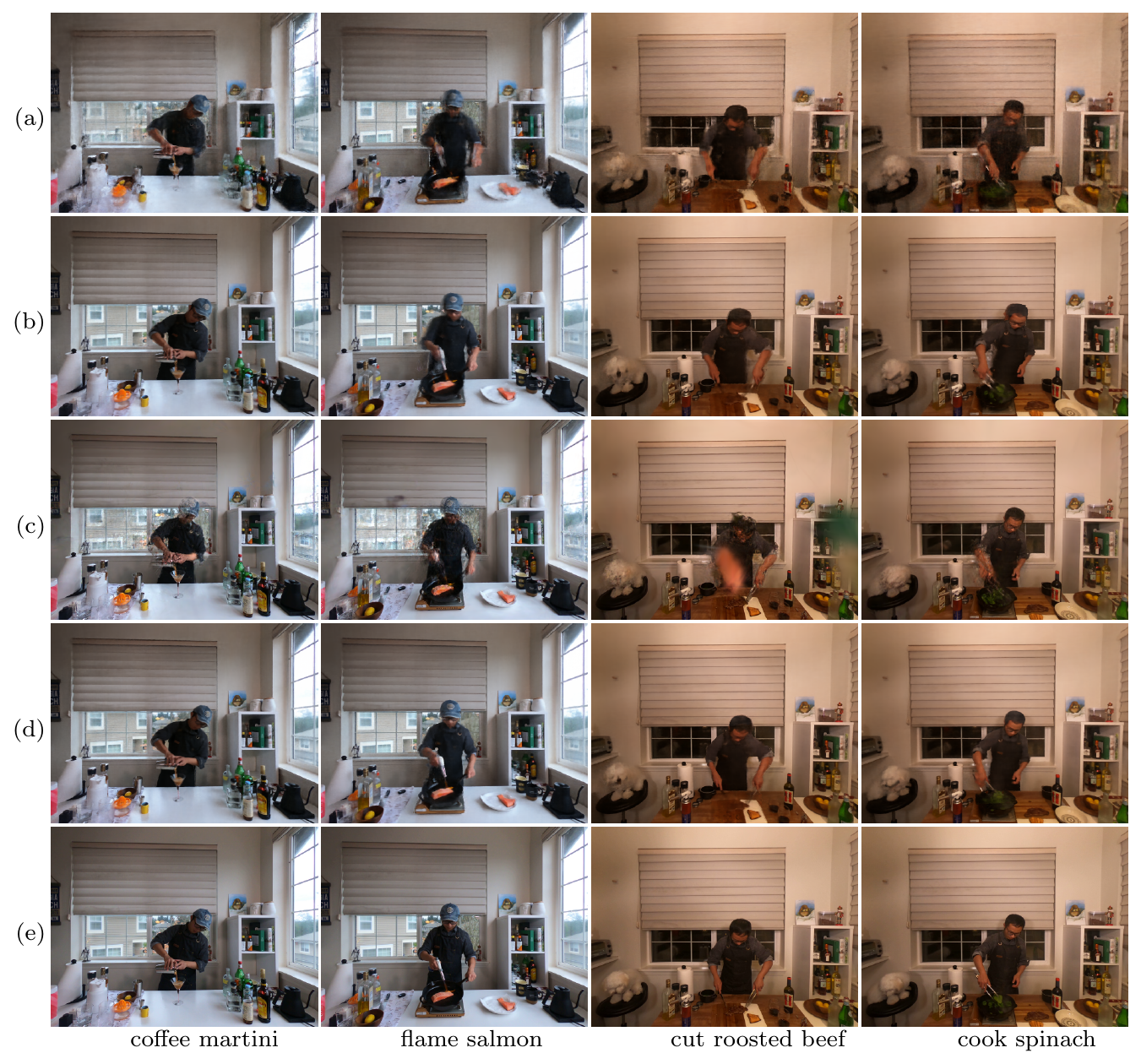} \vspace{-0.4cm}
\caption{ Visual comparison of different methods on Dynamic Scene Dataset. (a) Li \textit{et al.} \cite{li2021neural}, (b) Gao \textit{et al.} \cite{gao2021dynamic}, (c)Yang \textit{et al.} \cite{yang2023deformable} (d) Ours, (e) Ground Truth.}
\label{figure: result_3dvideo}
\end{figure*}

\noindent \textbf{Quantitative comparisons.}
Table \ref{tabel:results}  lists the PSNR, SSIM, and LPIPS \cite{zhang2018unreasonable} values of different methods on each of the scenes in the Dynamic Scene Dataset. From Table \ref{tabel:results}, it can be observed that  
\begin{compactitem}
    \item[$\bullet$] \textit{Ours} outperforms all comparison methods in terms of the average PSNR for 7 scenes and achieves the highest PSNR on 6 out of 7 scenes. Regarding SSIM and LPIPS, \textit{Ours} performs significantly better than other methods, except Gao \textit{et al.} \cite{gao2021dynamic} that utilizes pre-processed depth and optical flow maps predicted by well-trained state-of-the-art models, providing the network with more information and indicating the movement of objects at a pixel level. Although our model can regularize the motion part using unsupervised losses, this is much less straight and forceful compared to these explicit supervisions, which may result in blurry moving objects, thus limiting LPIPS;
    \item[$\bullet$] the superiority of \textit{Ours} in handling dynamic scenes is more evident in scenes with significant motion, such as Skating, Playground, and Umbrella, where it improves  the reconstruction quality of the second best method by more than 0.5 dB;
    \item[$\bullet$] \textit{Ours} outperforms the two NSFF-based methods, Li \textit{et al.} \cite{li2021neural} and Gao \textit{et al.} \cite{gao2021dynamic}, in PSNR on 6 out of 7 scenes. Although we adopt the same NSFF formulation as Li \textit{et al.}, their approach models the entire scene, including both static and dynamic parts, using a single formulation, which makes it difficult to maintain the structure of the static parts. We follow Gao \textit{et al.} in modeling the static and dynamic parts separately, which ensures the rigidity of the static parts. Additionally, while we do not use re-processed supervision, the comparison with Gao \textit{et al.}, who do, demonstrates the efficiency and superiority of our two proposed unsupervised constraints. 
    \item[$\bullet$] regarding the scene named Jumping, \textit{Ours} does not perform as well as Li \textit{et al}. \cite{li2021neural} and Gao \textit{et al}.  \cite{gao2021dynamic}. 
    The scene contains dynamic content primarily consisting of people jumping, which involves intricate non-rigid deformations. Our surface consistency constraint becomes invalid when dealing with non-rigid deformations.
    
    \item[$\bullet$]
    \textit{Ours} improves Tretschk \textit{et al.}~\cite{tretschk2021non} by 6.7 dB. Based on deformable NeRF, Tretschk \textit{et al.}~\cite{tretschk2021non} optimized a warp field that maps the canonical NeRF into a deformed scene at other times. Given that our experimental setting involves monocular videos with large motions, their method struggled to model complex motions throughout the sequence, as all frames are warped from the same canonical space. In contrast, the surface consistency constraint in our method is computed frame to frame, thus avoiding the challenges associated with large motions.
    \item[$\bullet$] particularly, \textit{Ours} significantly outperforms the most recent 3D-GS based method, Yang \textit{et al.} \cite{yang2023deformable}, across three metrics, notably achieving a superior PSNR by 3.58dB. Yang’s approach, which relies on 3D-GS and utilizes a direct point cloud representation, faces challenges in reconstructing complex scenes that require a large number of points for an accurate representation, especially when provided with insufficient image inputs. Our experimental setup, focusing on the use of monocular short videos, naturally limits the quantity of training data. This constraint adversely affects the efficiency of such methods.
\end{compactitem}

Table \ref{tabel:results_3dv} presents the PSNR, SSIM, and LPIPS \cite{zhang2018unreasonable} metrics for various methods applied to each scene within the Neural 3D Video Synthesis Dataset. The results in Table \ref{tabel:results_3dv} reveal that our method surpasses all others in PSNR, SSIM, and LPIPS across all six scenes, notably achieving a PSNR that is 1.5 dB higher than competing methods. The enhanced performance on the Neural 3D Video Synthesis Dataset can be attributed to the inadequacy of the pre-predicted flow/depth data, which does not suit the specific requirements of this dataset. Our method benefits from the use of implicit unsupervised loss terms, which prove to be more effective under these circumstances. This underscores the superior performance and effectiveness of our approach.

Note that in our framework, we adopt the \textbf{same} model as Gao \textit{et al}. \cite{gao2021dynamic} and Li \textit{et al}. \cite{li2021neural} to synthesize the  static background of a scene, and the proposed two constraints are \textbf{only} imposed on NSFF. 
To directly demonstrate the advantages of our constraints, we present the quantitative results of different methods \textbf{only on the dynamic part }of a scene in Table~\ref{tabel:results_d} and Table~\ref{tabel:results_3dv_dy}, where it can be seen that the superiority of our method over other methods is more significant.\\

\noindent \textbf{Qualitative comparisons.}
In Fig.~\ref{figure: result}, we compared the visual results of different methods on four scenes from the Dynamic Scene Dataset.
Our method produces better reconstruction results for dynamic objects compared to all other compared methods under the same experimental configuration. Yoon \textit{et al.} \cite{yoon2020novel} shows slight bias in the orientation and position of dynamic objects in complex scenes. Trestchk \textit{et al.} \cite{tretschk2021non} struggles to guarantee the integrity of synthetic objects in complex scenes and fails to model motion correctly for scenes with significant movements. While Li \textit{et al.} \cite{li2021neural} can accurately model dynamic objects, it produces blurry and less detailed results. Gao \textit{et al.} \cite{gao2021dynamic} fails to handle thin edges, such as the rope of the balloon and the strip of the umbrella, and generates blurry or artifact-laden regions containing large motions.  Yang \textit{et al.} \cite{yang2023deformable} fails to capture the whole dynamic part of the scene. We also refer the reviewers to the associated \textit{video demo}. Figure~\ref{figure: result_3dvideo} showcases a visual comparison of the reconstruction outcomes from various methods on four scenes within the Neural 3D Video Synthesis Dataset. Our approach also yields superior results in reconstructing dynamic objects over all other methods tested under identical experimental settings. The results from Li et al. \cite{li2021neural} appear blurry, those from Gao et al. \cite{gao2021dynamic} exhibit artifacts, and Yang et al. \cite{yang2023deformable} still do not fully capture the dynamic aspects of the scene.

\begin{table}[t]
\centering
\begin{center}
\caption{Quantitative results of the ablative study on the two proposed constraints. 
}\vspace{-0.4cm}
\label{table:ablation}
 \resizebox{0.5\textwidth}{!}{
\begin{tabular}{l|ccc}
\toprule[1.5pt]
  &PSNR$\uparrow$ & SSIM$\uparrow$ & LPIPS$\downarrow$  \\\hline
 Baseline & 24.81 & 0.822 & 0.161\\ 
 Baseline+$\mathcal{L}_{surface}$ & 25.95 & 0.838 & 0.154\\ 
 Baseline+$\mathcal{L}_{patch}$ &25.35 & 0.829 & \textbf{0.151}\\ 
 Full (Baseline+$\mathcal{L}_{surface}$+$\mathcal{L}_{patch}$) &\textbf{26.36} & \textbf{0.840} & \textbf{0.151}
\\
\bottomrule[1.5pt]
\end{tabular}
}
\end{center}
\end{table}

\begin{figure}[t]
\centering
\includegraphics[width=0.4\textwidth]{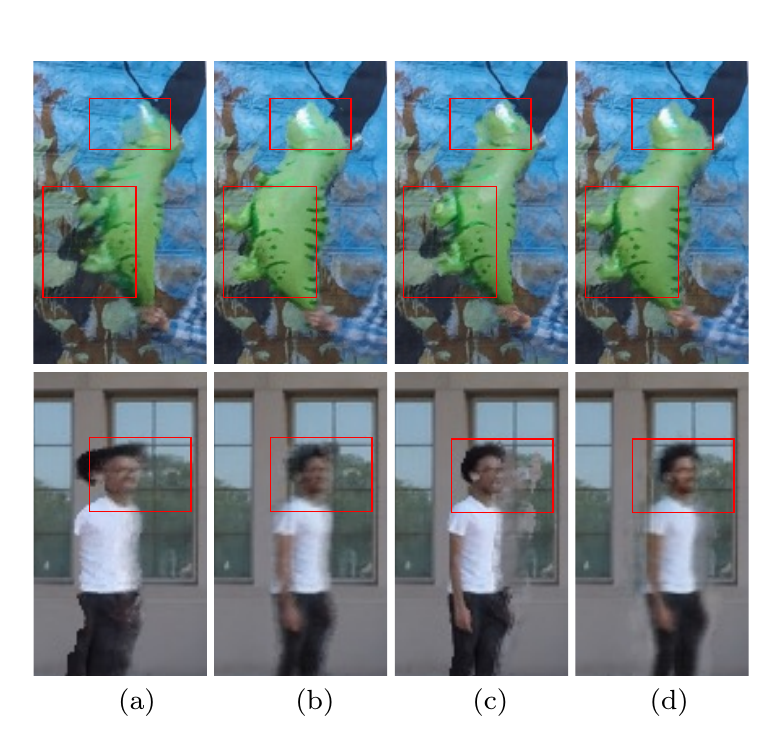}\vspace{-0.4cm}
\caption{Visual comparison of ablations of (a) Baseline, (b) Baseline+$\mathcal{L}_{surface}$, (c) Baseline+$\mathcal{L}_{patch}$, (d) Full model (Baseline+$\mathcal{L}_{surface}$+$\mathcal{L}_{patch}$). }
\label{figure: ablation}
\end{figure}

\subsection{Ablation Study}

We conducted ablation studies 
to understand the two proposed regularization terms better. 
We excluded these two terms from the complete framework to construct the baseline while keeping other regularization terms unchanged. 
Note that this baseline is actually the approach of Gao \textit{et al.} \cite{gao2021dynamic} without flow/depth supervision. To isolate the effects of the two key regularization terms on the dynamic objects, we used the same static model for all experiments in the ablation study, as we separated the static part and the dynamic part of scenes in our approach.

Table~\ref{table:ablation} lists the quantitative results, where it can be seen that each of the two terms can significantly improve the baseline, and the reconstruction quality is further improved largely when simultaneously using the two terms.  
Besides, as visualized in Fig.~\ref{figure: ablation}, the baseline model cannot predict the correct trajectory for dynamic objects, leading to distortions, artifacts, and blurs. The surface consistency constraint enables us to map the dynamic motions accurately and ensures object geometry consistency, while the patch-based multi-view consistency constraint produces sharper, clearer reconstructions with more details. The full model can accurately model dynamic movements and produce clear results, even for videos with large object motion.

\begin{figure}[t]
\centering
\includegraphics[width=0.5\textwidth]{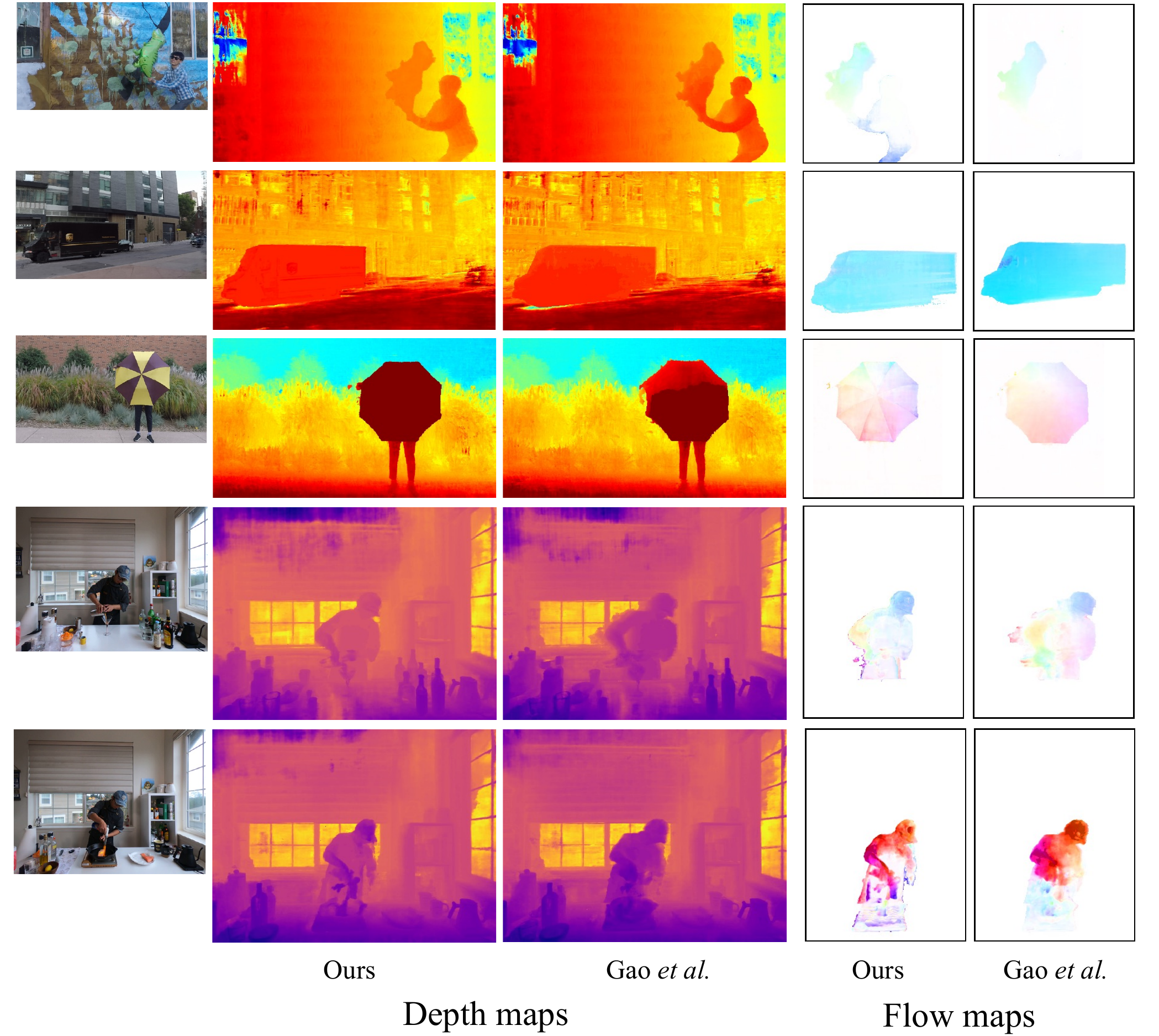} \vspace{-0.4cm}
\caption{Visual comparison of rendered depth maps and 
2D optical flow maps of our method and Gao \textit{et al.} \cite{gao2021dynamic}. 
We masked the optical flow maps using the estimated dynamic mask. 
}
\label{figure: flow_depth}
\end{figure}
\begin{table}[t]
\centering
\begin{center}
\caption{Comparison of the predicted scene flows by Gao \textit{et al.} \cite{gao2021dynamic} and Ours using PCK-T. $\uparrow$ means the larger, the better.
} \vspace{-0.4cm}
\label{table:pckt}
\resizebox{0.5\textwidth}{!}{
\begin{tabular}{c|c|c|c|c}
\bottomrule[1.5pt]
PCK-T$\uparrow$ &Balloon1&Balloon2&Jumping & Playground \\ \hline
 Gao \textit{et al}. \cite{gao2021dynamic}& \textbf{0.819}&0.886&\textbf{0.955}&0.962 \\
 Ours       &0.673&\textbf{0.909}&0.946&\textbf{0.970}
 \\ \hline \hline
 
PCK-T$\uparrow$ &Skating& Truck & Umbrella&-  \\ \hline
 Gao \textit{et al}. \cite{gao2021dynamic}&\textbf{1.00}&0.936&0.744&- \\
 Ours &\textbf{1.00 }&\textbf{0.978}& \textbf{0.767}&-
\\
\bottomrule[1.5pt]
\end{tabular}
}
\end{center} 
\end{table}

\subsection{Comparison of Flow and Depth Estimation}

Our method can predict the depth and flow maps of novel views in an unsupervised manner. Note that the 2D optical flow is obtained by projecting the predicted 3D scene flow to the camera space. 
For further analysis, as shown in Fig. \ref{figure: flow_depth}, despite some noise, our results exhibit strong structure awareness and can smoothly handle slight depth/flow changes across different parts of the same object. Particularly with the Neural 3D Video Synthesis Dataset, our approach more precisely retains the silhouette of the human figure in the flow/depth maps amidst casual movements, in contrast to Gao's method, which is prone to a greater incidence of artifacts. The flow and depth maps of novel views by our method are comparable to or even more accurate than those by 
Gao \textit{et al.} \cite{gao2021dynamic} that adopts pre-processed depth and flow maps as supervision, which is credited to our fine-grained formulation. 
To quantitatively evaluate the predicted scene flow, we utilized the 
PCK-T metric \cite{gao2022monocular}, which assesses 2D correspondences across frames by measuring the percentage of correctly transferred keypoints.
As listed in Table~\ref{table:pckt}, our method outperforms Gao \textit{et al.} \cite{gao2021dynamic} on 5 out of 7 scenes, demonstrating the efficiency of our unsupervised regularization terms. 
Since  most pixels of the moving object in scene Balloon1 have an identical color, equivalent to a textureless region, our flow estimation only driven by the unsupervised photometric loss is inaccurate.

\section{Conclusion and Discussion}
\label{sec: conclusion and discussion}

We have presented a new approach for dynamic view synthesis from a monocular video. Our method decouples the scene motion into object and camera motion and regularizes them separately with two loss terms: surface consistency constraint and path-based multi-view constraint, thus eliminating the need for pre-processed optical flow or depth maps as supervision. Impressively, our unsupervised method
even substantially improves  state-of-the-art supervised methods. Besides, our method can output more accurate scene flow and depth. We believe our method will provide a promising baseline for dynamic view synthesis.

Despite the encouraging performance, our method also suffers from a limitation, i.e., it is not proficient in addressing non-rigid deformation. The reason is that our approach relies heavily on the surface consistency constraint to model object movement. Regrettably, the surface consistency constraint becomes invalid when non-rigid deformation occurs. 
Besides, the general framework of existing works, including ours,  still requires separate modeling of the static and dynamic parts of the scene using two NeRFs and mask supervision for predicting which parts belong to the dynamic region. Inspired by the encouraging performance of our method, it is promising to simplify the modeling process even further to eliminate the need for such pre-processing steps and make the modeling process more efficient in the future. Furthermore, since our method aligns with established frameworks like NeRF and NSFF that prioritize forward-facing data processing, it may not perform optimally on casually captured videos, such as selfie videos. Therefore, it is a promising direction for future research to explore how to refine our approach and make it applicable to a wider range of casual and realistic videos. Additionally, it is worth noting that our method, being based on NeRF, can be relatively time-consuming. Therefore, exploring the integration of performance-improving and acceleration technologies into dynamic NeRF holds significant promise for future research endeavors.

Last but not least, while we replace pre-processed supervision information with two unsupervised constraints, it is worth noting that incorporating pre-processed supervisions such as depth and optical flow maps into the network can provide prior knowledge about scene geometry and motion, which can have a positive impact. In situations where there is a need to speed up or stabilize the training process, explicit supervision can also be considered as an option for our method.

{\small
\bibliographystyle{ieee_fullname}
\bibliography{egbib}
}

\end{document}